\documentclass{article}

\usepackage[final]{drlw_neurips_2019}

\usepackage[utf8]{inputenc} % allow utf-8 input
\usepackage[T1]{fontenc}    % use 8-bit T1 fonts
\usepackage{hyperref}       % hyperlinks
\usepackage{url}            % simple URL typesetting
\usepackage{booktabs}       % professional-quality tables
\usepackage{amsfonts}       % blackboard math symbols
\usepackage{nicefrac}       % compact symbols for 1/2, etc.
\usepackage{microtype}      % microtypography
\usepackage{amssymb}
\usepackage{color}
\usepackage{amsmath}
\usepackage{algorithm,algorithmic}
\usepackage{subcaption,graphicx}
\usepackage{wrapfig}

\newcommand{\Tau}{\mathcal{T}}

\title{Characterizing Policy Divergence for \\Personalized Meta-Reinforcement Learning}

\author{
  Michael Zhang\\
  Harvard University \\
  Cambridge, MA 02138\\
  \texttt{michael\_zhang@college.harvard.edu}
}

\begin{document}

\maketitle

\begin{abstract}
Despite ample motivation from costly exploration and limited trajectory data, rapidly adapting to new environments with few-shot reinforcement learning (RL) can remain a challenging task, especially with respect to personalized settings. Here, we consider the problem of recommending optimal policies to a set of multiple entities each with potentially different characteristics, such that individual entities may parameterize distinct environments with unique transition dynamics. Inspired by existing literature in meta-learning, we extend previous work by focusing on the notion that certain environments are more similar to each other than others in personalized settings, and propose a model-free meta-learning algorithm that prioritizes past experiences by relevance during gradient-based adaptation. Our algorithm involves characterizing past policy divergence through methods in inverse reinforcement learning, and we illustrate how such metrics are able to effectively distinguish past policy parameters by the environment they were deployed in, leading to more effective fast adaptation during test time. To study personalization more effectively we introduce a navigation testbed to specifically incorporate environment diversity across training episodes, and demonstrate that our approach outperforms meta-learning alternatives with respect to few-shot reinforcement learning in personalized settings.
\end{abstract}

% \begin{abstract}
%   The abstract paragraph should be indented \nicefrac{1}{2}~inch (3~picas) on
%   both the left- and right-hand margins. Use 10~point type, with a vertical
%   spacing (leading) of 11~points.  The word \textbf{Abstract} must be centered,
%   bold, and in point size 12. Two line spaces precede the abstract. The abstract
%   must be limited to one paragraph.
% \end{abstract}

% \MZ{What are the concrete advantages of model-free approaches here? Literature suggests model-based are more sample-efficient, but harder to model and maybe infeasible in naive consideration wrt potentially unbounded number of MDPs. However there is previous work that models each task as a timestep within an online-learning ``episode''}

\section{Introduction}
While reinforcement learning (RL) has demonstrated success in various sequential decision making settings \cite{barto1990sequential, shortreed2011informing}, challenges remain in the personalized domain. For example, we may be interested in recommending treatments to a cohort of patients, where individuals may respond to the same treatment differently, even while conditioning on the same health state. From an RL framework, each patient can be modeled as an entity whose individual characteristics parameterize a distinct Markov decision process (MDP). With this framing, one could, in principle, train a separate policy from scratch for each individual. However, this approach is impractical in settings where exploration is costly and individual trajectory data are limited. As a result, there is an increasing need to think about how to leverage prior data in sophisticated manners to develop few-shot learning approaches for deployment in unseen environments. 

In this work, we take a meta-learning approach to this problem and propose a method tailored to few-shot learning across a \emph{diverse} set of environments. Our approach builds on model agnostic meta-learning (MAML), a paradigm for learning a single prior initialization of policy gradient parameters that can adapt quickly to a set of observed tasks. We extend this approach by leveraging the notion that certain environments are more similar to each other than others in personalized settings, and accordingly their corresponding experiences should be given higher attention during adaptation. We thus propose generalizing this approach by learning several different potential initializations, and choosing an appropriate initialization to use for a given environment at test time.

% \newpage 

Our proposed algorithm--cluster adaptive meta-learning (CAML)--explicitly seeks to determine which past parameters should be factored in during adaptation. Because the core idea of our adaptation relies on being able to quickly identify the most relevant past trajectories, we seek to characterize these experiences through a distance metric that establishes a notion of similarity between policies, borrowing from related work in inverse reinforcement learning (IRL) and determining policy divergence. We rely on whole trajectories$-$which provide a richer source of information over alternatives used in previous work \cite{DBLP:journals/corr/abs-1803-11347, DBLP:journals/corr/FinnAL17, DBLP:journals/corr/abs-1812-07671}$-$and would like to call upon a diverse set of environment interactions for reference. Accordingly, we retain an episodic model-free approach in contrast to more recent model-based online-learning methods where each timestep is considered a new task \cite{DBLP:journals/corr/abs-1803-11347}. Our contributions center around how to (1) study personalization effectively in simulated contexts, (2) characterize the divergence of policies across a population of environments, and (3) learn effectively across different and previously unseen environments with minimum exploration cost. We introduce a $K$-medoids-inspired algorithm for policy adaptation and few-shot learning across multiple environments, and demonstrate approach's successes over existing methods in a 2D testbed.

\section{Background}

\subsection{Personalized Markov decision processes (MDPs)}
We consider a modified version of a typical continuous state-space MDP, parameterized specifically by an individual entity type $\Tau_i$ among a population of possible types $\Tau$. For lexical consistency, we refer to the agent as the actual policy or decision-maker, and define an ``entity'' to be the object through which an agent interacts with the larger environment (the ``patient'' in our working example). Each entity type then introduces an MDP  $\mathcal{M}^i \triangleq (\mathcal{S}, \mathcal{A}, P^i, r^i, \gamma, s_0^i)$ with unique transition probabilities $P^i: \mathcal{S} \times \mathcal{A} \times \mathcal{S} \rightarrow \mathbb{R}$. Given shared state and action spaces $S$ and $A$, we thus allow for diversity in behavior across entities from different types.

For any two different entity types $\Tau_i \neq \Tau_j$, $P^i$ may not necessarily differ from $P^j$, but crucially at the onset of training for some subset of unseen types $\Tau_i \subseteq \Tau$ we make no assumption that these transition dynamics are the same. Finally, $\gamma$ is a discount factor assumed to be the same for all agents, and $s_0^i$ is the initial state distribution for environment type $i$. Our goal across a population of entity types is then to learn some function $f : \Tau \rightarrow \Pi$ mapping from entity types to policies. For each entity type given $s^i_0$ and a minimal number of timesteps $t$, we can then output an optimal policy $\pi_\theta^i$ with regard to maximizing the cumulative discounted reward $\sum_{t=0}^{T - 1} \gamma^t r(s_t, a_t)$ for $T$-length episodes, where $\theta$ denotes the parameters for personalized policy $\pi_i$ taking action $a_t \in \mathcal{A}$ given state $s_t \in \mathcal{S}$.

\subsection{Meta-reinforcement learning}
In the typical meta-learning scenario, we are interested in automatically learning learning algorithms that are more efficient and effective than learning from scratch \cite{DBLP:journals/corr/abs-1803-11347}. To do so, instead of looking at individual data points, a meta-learning model is trained on a set of tasks, treating entire tasks as training examples with the goal being to quickly adapt to new tasks using only a small number of examples at test time. We can consider a task to be any type of desirable machine learning activity, such as correctly classifying cat images in a supervised learning scenario, or learning how to walk forwards at a certain velocity in RL. Expanding on this latter example, meta-RL tries to learn a policy for new tasks using only a small amount of experience in the test setting. More formally, each RL task $\tau_i$ contains an initial state distribution $s_0^i$ and transition distribution $P^i(s_{t+1} | s_t, a_t)$. In the few-shot RL setting parameterized by $K$ shots, after training on train tasks $\{\tau_\text{train}\}$, we are allowed to use $K$ rollouts on test task $\tau_\text{test}$, trajectory $(s_0, a_0, \ldots s_T)$, and rewards $R(s_t, a_t)$ for adapting to $\tau_\text{test}$.

\subsection{Model-agnostic meta-learning for RL}

More specifically, our approach builds on model-agnostic meta-learning (MAML) for reinforcement learning \cite{DBLP:journals/corr/FinnAL17}, which optimizes over the average reward of multiple rollouts. Doing so we wish to learn an initial set of policy gradient parameters capable of quickly adapting to a new RL task in a few gradient descent steps. Given $n \in N$ meta-training environments, and initial policy parameters $\theta$, MAML first samples an MDP $\mathcal{M}^i$, $i \in \{1, 2, \ldots, n\}$. For each sampled MDP, we then train a policy given $K$ rollouts, arriving at task-adapted parameters $\theta_i'$ through gradient descent $\theta_i' = \theta - \alpha \nabla_\theta \mathcal{L}_{\mathcal{T}_i}(\pi_\theta)$ (described as the inner training loop). $\alpha$ is our learning rate and $\mathcal{L}_{\mathcal{T}_i}(\pi)$ describes the inverse reward from following policy $\pi$ with entity type $\mathcal{T}_i$. 
Feedback through the reward generated on the $(K+1)^{\text{st}}$ rollout is then saved. After $n$ tasks are sampled, we perform a meta-update using each saved reward, updating our initialization parameters $\theta'$ with $\theta' \leftarrow \theta - \beta \nabla_\theta \sum_{\Tau_i \sim p(\Tau)} \mathcal{L}_{\Tau_i}(\pi_{\theta_i'})$, where $\beta$ is the meta-learning rate. MAML thus seeks to leverage the gradient updates for each task during the inner loop of training, reaching a point in parameter space relatively close to the optimal parameters of all potential tasks. We then evaluate performance by sampling $\mathcal{M}^j$, $j \in N \setminus n$, observing the rewards generated from the $(K+1)^{\text{st}}$ trajectory after adapting with $K$ rollouts and policy parameters $\theta'$.

\subsection{Characterizing policy divergence}

Aside from gaining further knowledge with respect to how policies adapt to their environments over time, we also seek to establish a valid distance metric between various policies to inform policy parameter initialization in few-shot RL. Conceptually we can imagine that given pairs of MDPs and policies $(\mathcal{M}^i, \pi^i)$ for types $\Tau_i \in \Tau$, the policies start with an indistinguishable parameter initialization and diverge when adapting to their personalized environments after $t$ timesteps of training. If adapted policies $\pi_i'$ and $\pi_j'$ are still similar to each other, then this suggests that their environments $\mathcal{M}^i$ and $\mathcal{M}^j$ may be similar as well, and accordingly we can use this information to form a representation of optimal policy parameter clusters, making sure to only reference relevant prior experiences when adapting to a new environment.

To begin our search for such a distance metric we look to the field of imitation learning, and specifically inverse reinforcement learning (IRL). Here the goal is to learn a cost function explaining an expert policy's behavior \cite{Ng:2000:AIR:645529.657801}. However, while typical IRL methods first learn this cost function and then optimize against it to train a policy to imitate the expert, we are content with the cost function alone, whose components we can borrow to define our own quantifiable metric defining the distance between two policies. As noted in \cite{DBLP:journals/corr/HoE16}, we can uniquely characterize a policy by its occupancy measure given by 
\begin{equation}
\rho_\pi(s, a) = \pi(a |s) \sum_{t= 1}^{T} P(S_t = s | \pi), \text{ where } \rho_\pi : \mathcal{S} \times \mathcal{A} \rightarrow \mathbb{R}
\end{equation}
We essentially view this metric as  the state-action distribution for a policy. Using a symmetric measure such as the Jensen-Shannon (JS) divergence \cite{lin1991divergence}, we can thus compare the occupancy measures of multiple policies at specific time steps over training as a metric into policy divergence over time. While we study a very different problem from traditional IRL–where we do not explicitly try to minimize policy divergence–such measures of divergence allow for both descriptive analysis in tracking divergence and informing new learning algorithms. In principle we can then characterize policies based on their observed trajectories, and later show that this is enough to measure similarities across their respective environments as well.

\section{Related Work}
While there is much work summarizing the general meta-learning literature \cite{DBLP:journals/corr/abs-1803-11347,DBLP:journals/corr/abs-1810-02334, DBLP:journals/corr/abs-1803-00676}, we focus on initialization-based methods for RL. Here we wish to learn a set of optimal model parameters such that after meta-learning, the policy is initialized to an optimal position in parameter space to adapt to new environments relatively quickly. Finn et al. demonstrate this paradigm in MAML, which achieves meta-reinforcement learning through gradient-based optimization \cite{DBLP:journals/corr/FinnAL17}. They introduce the generalizable notion of doing well across a variety of tasks, where each task in the RL setting is a similar goal such as trying to reach a certain point in 2D space. Given training on reaching a certain support set of points, we would like to be able to quickly learn how to reach a new batch of unseen query points. Instead of trying to generalize across strictly different tasks, we adopt this framework by modeling different entity types encountered$-$each harboring their own potentially unique rewards and transition probabilities given the same states and actions$-$as different tasks. 

Implementation-wise, although modern day software packages make MAML's computation relatively straightforward, it's reliance on unrolling the computation graph behind gradient descent and taking second derivatives has motivated follow-on work on simpler methods, such as first-order MAML and Reptile, which avoid doing so with first-order approximations and direct gradient movements respectively \cite{DBLP:journals/corr/FinnAL17, DBLP:journals/corr/abs-1803-02999}. We focus on learning algorithms with similar computational dependencies due to their comparable performance with vanilla MAML.

Learning similarities between tasks through clustering or other distance-based  metrics has also been explored before in non-RL few-shot settings, where a central motivation lies in being able to achieve comparable results with alternatives while maintaining much simpler inductive biases. Snell et al. propose prototypical networks for few-shot classification \cite{DBLP:journals/corr/SnellSZ17}, which aims to learn an embedding where data points cluster around a single prototype representation in each class. Accordingly, given a learned transformation between input features and some vector space, we can simply use the computed prototypes on unseen data points for classification. Although we do not explore learning further embeddings on our proposed distance metric, we nonetheless demonstrate that calculating similarities given observed occupancy measures is enough to identify similar environments, drawing upon a similar intuition for RL.

\section{Personalizing policy learning}
% Additionally, with respect to scenarios in mHealth, we share the revised meta-RL setup presented in \cite{DBLP:journals/corr/abs-1803-11347}, where for a continuous rather than episodic test environment, $K$ corresponds to individual timesteps $t$ as opposed to entire trajectories. 

% In terms of actual learning, we reference model-based RL, which aims to solve the classical MDP optimization problem by explicitly learning a transition distribution $P(s' | s, a)$. In our mHealth setting where exploration is costly, we wish to  estimate $\hat{p}_\theta(s'|s, a)$ to approximate the dynamics of an environment produced by an agent type $i$. Traditionally, $\theta$ is optimized to maximize the log-likelihood of observed trajectory data $\mathcal{D}$. Our goal will then be to find $\theta^{t*} = \arg \max_\theta \mathcal{D}^{t}$ for test agent type $t$.
We now present our approach for (1) measuring and interpreting divergence of policies in personalization, and (2) defining few-shot meta-learning algorithms for online adaptation to new entity types. We also introduce a testbed to study the effects of personalization across entities on training. As outlined in Section 2.1, we work with a set of personalized MDPs representing a population of entity types.
% We also introduce a testbed (with more information in the Appendix)
% to study the effects of personalization across entities on training. As outlined in Section 2.1, we work with a set of personalized MDPs representing a population of entity types. 

% We expand upon our previous characterization of agent types $i$, by introducing trajectory segment $\tau^i(j, k)$, which represents a sequence of states and actions $\{s_j, a_j, \ldots, s_k, a_k, s_{k+1}\}$ sampled from the MDP $\mathcal{M}^i$. Letting $\tau^i(j, k) = \tau^i$ in general when convenient, we have $\tau^i \in \mathcal{D}^i$ for total data collected when interacting with agent type $i$. As our policies train, we can expect to amass large datasets $\mathcal{D}^i$ for each individual agent type in our training sample. Furthermore, given the differences in agents, we may expect differences in optimal personalized policies $\pi^*_i$, and would like to characterize this divergence over time.

\subsection{A personalized particle environment for 2D navigation}
\begin{wrapfigure}{r}{0.5\textwidth}
%   \vspace{5.5pt}
  \vspace{-13.5pt}
  \begin{center}
    \includegraphics[width=0.48\textwidth]{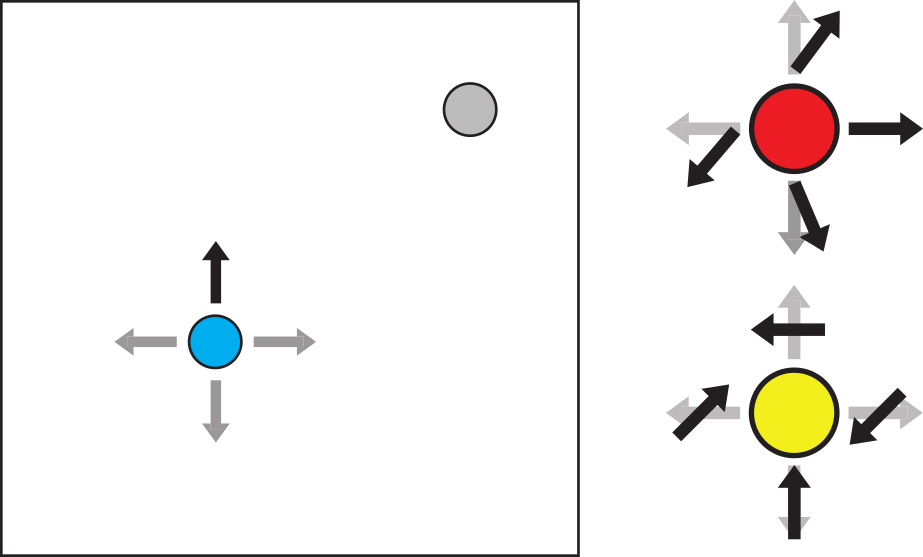}
  \end{center}
  
  \caption{\textbf{2D personalized particles.} Policy $\pi$ tries to move entity
 (blue, red, yellow) to target (gray), but entities behave in different ways unknown to the policy. Right: Two remapping schemes, where some cardinal direction e.g. `up' corresponds to a different transition vector. Code available upon request.}
 \vspace{-13.5pt}
\end{wrapfigure}
We begin our study of divergence in personalizing policies with a 2D continuous episodic gridworld environment, where a point agent must move to a target position in a constrained amount of time. The state space is given by the agent's current 2D position, and at each timestep the agent may choose to go left, right, up, or down one pixel unit. Following these commands, the agent tries to maximize its reward, calculated to be the negative squared distance to the goal after each timestep. Finally episodes end every $100$ timesteps, or when the agent is within $0.01$ units of the goal. Particles start at origin $(0, 0)$, and try to reach target at $(1, 1)$ before the episode ends. $\mathcal{S}$ is defined by the coordinates of the particle at timestep $t \in \{0, 1, \ldots T\}$. Actions are discrete $a$, corresponding to moving one unit left, right, down, up, i.e.
$a_t \in \{(-1, 0), (1, 0), (0, -1), (0, 1)\}$.

We personalize this environment with the introduction of a population of entities, where we may either stick with one throughout the policy's entire training run or randomly introduce new entities into the world at the start of every episode. Each agent is initialized with a personalization function $\mathcal{F}: \mathcal{A} \rightarrow \mathcal{A}'$, which first remaps the cardinal directions of each action and then imposes additional variance. For example, given default action $a_r = (1, 0)$, telling a default agent to go right, our policy may encounter an agent who upon receiving action input $(1, 0)$ actually moves with action $(0.42, -1)$. This is achieved by first remapping $a_r$ to $(0, -1)$, and then imposing further variance on the x-coordinate. We can think of this as having the same set of interventions for all agents, but facing different responses in the form of varied actual transition outcomes.

\subsection{Practical divergence estimation}
As introduced in Section 2, by characterizing policies through their occupancy measures, we derive a similarity metric and quantitatively analyze their divergence. Furthermore, individual policies may move closer or further to others over time, forming new neighbors that reflect personalization over time. However, as discussed in \citep{DBLP:journals/corr/HoE16}, simply using the occupancy measure as presented and matching on all states and actions is not practically useful. 
% One concern is the continuous nature of our state space.
Instead of looking at an occupancy measure defined over all states as above then, we propose an alternative measure reliant only on a sampled subset of states. We observe that two policies may intuitively be different if for each of various given states, they differ in their action distributions. However, the states must have a non-zero chance of occurring in both policies of interest. Accordingly, in the $K$-shot $N$-ways setting, where for each of $N$ MDPs $\mathcal{M}_i$ we get $K$ trajectories $\{\tau^1_i, \ldots, \tau^K_i\}$ to observe before updating and an additional trajectory $\tau'_i$ after the update, we can calculate a kernel density estimate (KDE) $\hat{q}(\tau_{1: N})$ based on the observed trajectories $\tau_{1: N}' = \{\tau'_1, \tau'_2, \ldots \tau'_N\}$. States $\hat{s}_j \sim \hat{q}(\tau_{1: N}')$ can then be sampled, with probability density $\hat{q}(\tau_{1: N}')(\hat{s}_j)$.

Referring back to (1), for any policy $\pi_i$, we estimate the conditional probability $\pi(\cdot | s)$ by feeding in $\hat{s}_j$ to our model network and obtaining the probability vector $\pi'_i(\hat{s}_j) = (p_1^j, \ldots, p_n^j)$ where $p^j_a$ denotes the probability of taking action $a$ given state $s_j$. Towards estimating $\sum_{t=1}^{T} P(S_t = s | \pi)$, we again calculate the KDE over our observed trajectories, but now only consider $\tau'_i$ for MDP $\mathcal{M}_i$. 

Accordingly, for any set of policies we first obtain a respective observed batch of trajectories, calculate an estimated state-marginalized variant of occupancy measure $\hat{\rho}_s^i$ for $s \in \mathcal{S}$ and entity types $\Tau_i$, and compute pairwise symmetric divergences across entity types. In our case we use the Jensen-Shannon divergence \cite{lin1991divergence}.
\begin{equation}
    D_{\text{JS}}(\rho^{i}_s, \rho^{j}_s) \triangleq D_{\text{KL}}\Big(\rho^{i}_s\; \big{|}\big{|}\; \frac{\rho^{i}_s + \rho^{j}_s}{2}\Big) + D_{\text{KL}}\Big(\rho^{j}_s\; \big{|}\big{|}\;\frac{\rho^{i}_s + \rho^{j}_s}{2}\Big) \text{ for } i \neq j
\end{equation}
where $D_\text{KL}(\cdot, \cdot)$ is the Kullback–Leibler divergence. 
As a final measure, we employ $K$-medoids clustering to group our policies using the pairwise $D_{\text{JS}}(\cdot, \cdot)$ as a distance metric. The overall procedure is summarized in Algorithm 1. 
Besides describing our policies, we use these metrics to inform algorithms for few-shot learning, as described in the next section.

% Hsu et al. (2018) show that even simple clustering algorithms (e.g. k-means clustering) on an embedding space allows for downstream improvement in complicated tasks with respect to the original representation \cite{DBLP:journals/corr/abs-1810-02334}. While their results are derived from four image classification datasets, here we propose clustering on RL.

% From model adaptation perspective, clustering reduces search space? 

% In general try to find the transition model that maximizes the likelihood of the observed data, and also the chance of a positive reward?

% \subsection{An algorithm to identify similar policies}

% Consider different intervals $\tau = [t, t + \delta t]$ to themselves be different tasks as in the standard transfer-RL literature? Accordingly we can apply method.

%
% \begin{minipage}[t]{0.46\textwidth}
%   \vspace{0pt}
\begin{algorithm}[b]
\caption{Policy Divergence Estimation through Observed Trajectories}
\label{alg1}
\begin{algorithmic}[1]
\REQUIRE Distribution over entity types $p(\Tau)$, initial policies $\pi^i_\theta \in \Pi$ and episode horizon $T$
\REQUIRE Initialize update counter $t = 0$, and batch size $n$ for computing estimates
% \STATE For each type $\Tau_i \in \Tau$, initialize policies $\pi^{i}_\theta$
\WHILE{not done}
    \STATE Sample entity type $\Tau_i \sim p(\Tau)$ and initialize new policy $\pi_\theta^i$
    \STATE Sample $K$ trajectories $\tau^i = \{s_1, a_1, \ldots, s_T\}$ following $\pi^i_\theta$
    \STATE Update $\theta'$ through vanilla policy gradient with REINFORCE \cite{NIPS1999_1713}
    \STATE Save trajectory $\tau'_i = \{s_1, a_1, \ldots, s_T\}$ using updated
    $\pi_{\theta'}^i$
    \STATE Increment $t \leftarrow t + 1$
    \IF{$t = n$}
        \STATE Compute type-specific KDE $\hat{q}(\tau_i')$ for all saved $\tau_i^{'}$
        \STATE Obtain concatenated updated trajectories $\tau_{1:N}' = \{\tau'_1, \ldots, \tau'_N\}$
        
        \STATE Compute overall KDE $\hat{q}(\tau_{1: N})$ and generate samples $s \sim \hat{q}(\tau_{1: N}) $
        \FORALL{$s$}
             \STATE Calculate $\pi^i(a | s)$ for all policies $\pi^i$ and compute state probability densities $\hat{q}(\tau_i')(s)$
             \STATE Estimate $\hat{\rho}_s^i = \pi^i(a | s)\hat{q}(\tau_i')(s)$ for given state $s$
        \ENDFOR
    \STATE Compute time-indexed pairwise Jensen-Shannon divergences
    $D_{\text{js}}(\hat{\rho}_s^i, \hat{\rho}_s^j)$
    for all $\Tau_i, \Tau_j \in \Tau$
    \STATE Obtain divergence metric $\sum_{s \sim \hat{q}(\tau_{1:N}')}D_{\text{js}}(\hat{\rho}_s^i, \hat{\rho}_s^j)\hat{q}(\tau_{1:N}')(s)$
    \STATE Reset $t = 0$
    \ENDIF
\ENDWHILE
\end{algorithmic}
\end{algorithm}
% \end{minipage}%
% \hfill
% \begin{minipage}[t]{0.46\textwidth}
%   \vspace{0pt}  

% \end{minipage}

% \newpage

% \subsection{Meta-RL through $\boldsymbol{K}$-medoids clustering}
\subsection{Cluster-adapting meta-learning}
Given our policy divergence estimators, we hope to better organize our previous experiences for adaptation in new environments. In this section, we describe a $K$-medoids-inspired meta-learning algorithm called cluster-adapting meta-learning (CAML). Similar to serial versions of meta-learning algorithms such as MAML, CAML iteratively learns an initialization for parameters of a neural network model, such that given new MDP $\mathcal{M}^i$, after a few trajectory rollouts and a single batch update during test time, our policy performs competitively to those pretrained on the same environment. However, while previous algorithms update a single set of optimal parameters drawn from parameter space, CAML maintains parameters representative of larger clusters.

In setting up CAML, we consider the following design choices: (1) a valid distance metric to compare past trajectories and their corresponding policies in a model-free setting, (2) the clustering algorithm to organize past experiences, and (3) an initialization method to select the most appropriate policy parameters at test time. We consider (1) to be the most important contribution, and included details in Section 4.2. As described, we implement (2) using the $K$-medoids algorithm \cite{kaufmann1987clustering}, although in practice any clustering method may be used.
% Although additional work must be done to identify optimal initialization parameters for a given new environment, 
% We show that for $K = 10$ trajectory rollouts CAML performs favorably. As described, we implement clustering 
% \begin{figure}[h]
%   \centering
%   \begin{subfigure}{.3\textwidth}
%     \centering
%     \includegraphics[width=1.2\linewidth]{figures/checkpoints-0.png}
%     % \caption{Agent type 1}
%   \end{subfigure}%
%   \hfill
%   \begin{subfigure}{.3\textwidth}
%     \centering
%     \includegraphics[width=1.2\linewidth]{figures/checkpoints-20.png}
%     % \caption{Agent type 2}
%   \end{subfigure}%
%  \begin{subfigure}{.3\textwidth}
%     \centering
%     \includegraphics[width=1.2\linewidth]{figures/final_dest-136.png}
%     % \caption{Agent type 1}
%   \end{subfigure}%
%   \hfill
%   \begin{subfigure}{.3\textwidth}
%     \centering
%     \includegraphics[width=1.2\linewidth]{figures/final_dest-245.png}
%     % \caption{Agent type 2}
%   \end{subfigure}%
%   \caption{\textbf{Policy divergence over time.} T-SNE visualization of policy divergences over $40$ training updates. Colors denote original latent group. Computing pairwise distances using our divergence metric leads to noticeable grouping over time.}
%   \vspace{-12pt}
% \end{figure}
Finally with regard to $(3)$ at the end of our training iterations we obtain $k$ medoid policies, and during evaluation we view fast-adaptation to each individual MDP as a bandit problem. Given $k$ available arms (the policies) and $K$ arm pulls (the few number of shots, i.e. episodes allowed), we want to maximize the corresponding reward from following the arm policy for an entire episode. While more sophisticated multi-armed bandit methods exist \cite{DBLP:journals/corr/abs-1902-03657}, we found that simply sampling policies and saving their cumulative associated rewards for each rollout $r \in \{1, 2, \ldots, K\}-$then initializing with the medoid policy parameters corresponding to the highest reward$-$performed competitively against baselines. The full method is described in Algorithm 2. Following these considerations, we found that in few-shot settings ($K = 10$ trajectory rollouts) CAML performs favorably to alternatives.  
% The full method is described in Algorithm 2 in the appendix.

\begin{algorithm}[H]
\caption{Training with CAML ($K$-medoids version)}
\label{alg1}
\begin{algorithmic}[1]
\REQUIRE Distribution over entity types $p(\Tau)$, initial policy parameters $\theta$, episode length $T$
\REQUIRE Number of medoids $k$, batch size to cluster with $n$
% \STATE For each type $\Tau_i \in \Tau$, initialize policies $\pi^{i}_\theta$
\FOR{initial iterations $1, \ldots, n$}
    \STATE Sample entity type $\Tau_i \sim p(\Tau)$ and initialize new policy $\pi_\theta^i$
    \STATE Obtain and save updated $\pi^i_{\theta'}$ and $\tau_i'$ through Steps $3, 4, 5$ in Algorithm 1
\ENDFOR
\STATE Compute distance matrix $D$ with pairwise distances on the saved $\tau_i'$ using Algorithm 1. 
\STATE Perform $K$-medoids clustering on $D$, saving set of $k$ corresponding medoid policies $\Pi_\theta^k$ and trajectories $\tau^k_{1:N}$. Save trajectories $\tau^k \in \tau^k_{1:N}$ to compute next iteration of occupancy measures.
\FOR{iterations $n+1, n+2 \ldots$}
    \STATE Sample entity type $\Tau_i \sim p(\Tau)$ and randomly sample medoid policy $\pi_\theta \in \Pi^k$
    \STATE Obtain and save updated $\pi^i_{\theta'}$ and $\tau_i'$ through Steps $3, 4, 5$ in Algorithm 1
    \IF{number of saved policies equals $n$}
        \STATE Repeat steps $5$ and $6$, updating $k$ medoid policies and trajectories.
    \ENDIF
\ENDFOR
\end{algorithmic}
\end{algorithm}

% we randomly pick a policy and observe the reward, saving rewards for each policy as we go. At the end of $K$ rollouts, we then choose the medoid policy parameters corresponding to the highest max reward to initialize with.

\section{Experiments}
In our experimental evaluation, we validate if: (1) clustering on our occupancy measure estimate reasonably characterizes divergence of policies over time, (2) our environments suggest that personalization is necessary for fast adaptation, and (3) building on these metrics can inspire effective few-shot learning algorithms. To focus on the effect of our algorithm we use vanilla policy gradient (VPG) for both meta-learning and evaluation. 
% As previously described, we introduce a modified RL testbed inspired by classic 2D navigation \cite{DBLP:journals/corr/FinnAL17}, where we incorporate environment diversity 
% In this section we validate if meta-learning with occupancy measure clustering can inspire effective few-shot learning algorithms. We further include results on occupancy measure as a reasonable metric in the Appendix. To focus on the effect of our algorithm we use vanilla policy gradient (VPG) for both meta-learning and evaluation. We introduce a modified RL testbed inspired by classic 2D navigation \cite{DBLP:journals/corr/FinnAL17}, where we incorporate environment diversity 
% Details are in the appendix, but briefly, we introduce environmental diversity 
% through variance in the direction and speed associated with the same action across different entity types.

\subsection{Matching policy divergence to training environments}
Towards evaluating both the dynamics of our testbed and the strength of our policy divergence distance metric, we first sought to see if clustering on trained VPG trajectories could recover their corresponding initial environmental dynamics. To do so, we first initiated six latent entity types by remapping controls completely (e.g. the yellow particle in Figure 1). Afterwards, for each latent type, we generated four variants by further introducing uniform variance for given entity and action while preserving the defining cardinal direction (red particle in Figure 1). In total we ended up with $24$ entity types. To eludidate divergence across optimal policies, for each type we then trained an individual VPG policy from scratch over $40$ updates performing a batch parameter update every $10$ iterations, and calculating estimated occupancy-measure-based distances according to Algorithm 1.

\begin{figure}[t]
  \centering
  \begin{subfigure}{.33\columnwidth}
    \centering
    \includegraphics[width=\linewidth]{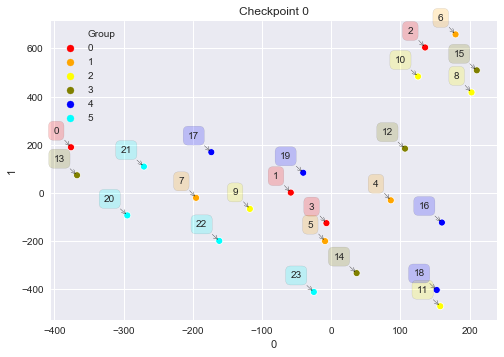}
    % \caption{Agent type 1}
  \end{subfigure}%
  \hfill
  \begin{subfigure}{.33\columnwidth}
    \centering
    \includegraphics[width=\linewidth]{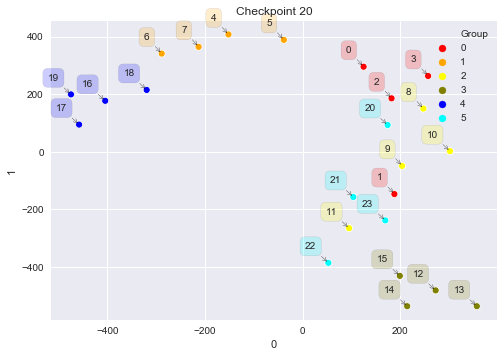}
    % \caption{Agent type 2}
  \end{subfigure}%
  \hfill
  \begin{subfigure}{.33\columnwidth}
    \centering
    \includegraphics[width=\linewidth]{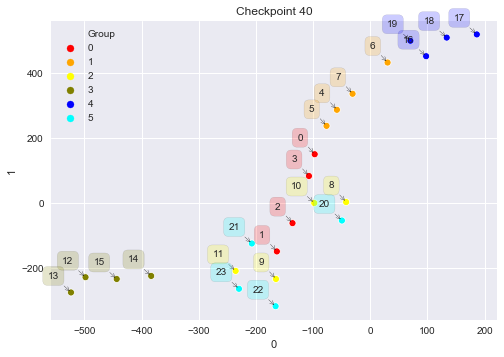}
    % \caption{All agent types}
  \end{subfigure}
  \caption{\textbf{Policy divergence over time.} T-SNE visualization of policy divergences over $40$ training updates. Colors denote original latent group. Computing pairwise distances using our divergence metric leads to noticeable grouping over time.}
  \vspace{-12pt}
\end{figure}

As observed in Figure $2$, based on estimated occupancy measure, trained policies show increasing signs of neighboring with members of their respective groups over training updates. One takeaway is that our occupancy measure-based divergence is effective at measuring policy divergence, where directed through their respective environments, we can thus evaluate how different two policies are based on their trajectories. % \begin{wrapfigure}{r}{0.33\textwidth}
% %   \vspace{5.5pt}
%   \vspace{-9pt}
%   \begin{center}
%     \includegraphics[width=0.33\textwidth]{figures/checkpoint-progress-0.png}}
%   \end{center}
%  \caption{Example policies improving with training over time}
%  \vspace{-12pt}
% \end{wrapfigure}
% Additionally, comparing the divergence calculations with training progress for example policies, the policies converging to optimal performance over time (Figure 3) suggest that expert policies trained on similar environments are close to each other in occupancy measure space. 
Accordingly, in the few-shot RL setting, one valid strategy for producing an optimal policy without having to train from scratch would be to identify prior expert policies trained on similar environments. Taking this one step further, because our occupancy measures directly derive from $\pi(a | s)$ which is itself derived from a policy network's parameters, this provides another interpretation into the effectiveness of finding nearby parameters in policy space. 

\subsection{Evaluating fast adaptation to personalized environments}

\begin{figure}[b]
  \centering
  \begin{subfigure}{.3\columnwidth}
    \centering
    \includegraphics[width=\linewidth]{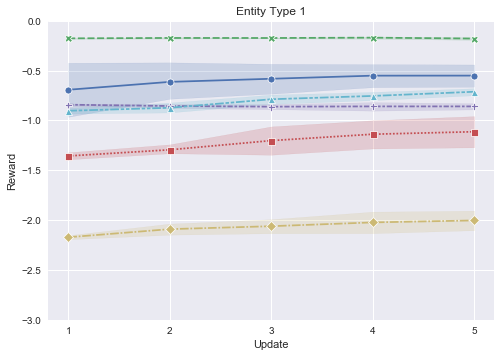}
    % \caption{Agent type 1}
  \end{subfigure}%
  \hfill
  \begin{subfigure}{.3\columnwidth}
    \centering
    \includegraphics[width=\linewidth]{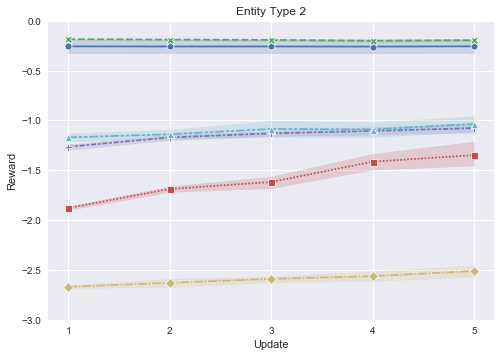}
    % \caption{Agent type 2}
  \end{subfigure}%
  \hfill
  \begin{subfigure}{.39\columnwidth}
    \centering
    \includegraphics[width=\linewidth]{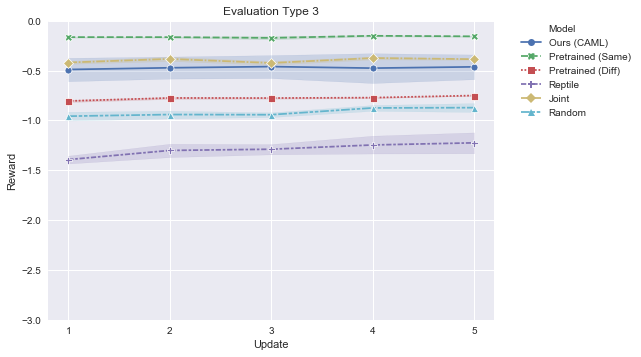}
    % \caption{All agent types}
  \end{subfigure}
  \caption{\textbf{Evaluation of learning algorithms on personalized environments.} CAML performs comparatively well across various evaluation entity types. }
%   \vspace{-30pt}
\end{figure}

% We also introduce a testbed (with more information in the Appendix)
% % to study the effects of personalization across entities on training. As outlined in Section 2.1, we work with a set of personalized MDPs representing a population of entity types. 

% We detail our testbed in the Appendix, but modify a classic 2D navigation setting  to introduce diversity in entity directions and speeds

% the population through  

Given training on a set support types $\Tau^S$, our learning algorithms should quickly perform well on unseen query type set $\Tau^Q$. In-line with $K$-shot RL, for each support type $\Tau_i^S \in \Tau^S$, our policies are allowed $K$ rollouts for adaptation during each training iteration. Towards evaluation, for each learning algorithm we train with $100$ iterations using its policy-specific training algorithm. All underlying gradients were calculated with VPG, updating with batch size $10$. For evaluation, due to the on-policy nature of policy gradient, we first allow $K$ rollouts to collect new samples from unseen type $\Tau_i^Q \in \Tau$. Allocating these trajectories as an initialization period, for each $\Tau_i^Q$ we then evaluate fast adaptation by comparing rewards across policies for up to five updates with fine-tuning via VPG.  

We compare CAML to five other methods: (1) pretraining a policy by randomly sampling from all support environments during training, similar to joint training \cite{DBLP:journals/corr/abs-1709-07979}, (2) pretraining a policy in the same query environment that we use for evaluation, (3) pretraining a policy in a different environment, (4) training with Reptile, a meta-learning algorithm comparable to MAML in performance but much simpler in computation that performs $K > 1$ steps of stochastic gradient descent for randomly sampled support environments before directly moving initialization weights in the direction of the weights obtained during SGD \cite{DBLP:journals/corr/abs-1803-02999}, and (5) randomly initializing weights. 

The results in Figure 3 suggest that CAML is able to quickly adapt to unseen entity types after one update during evaluation. We note that given three different environments, CAML consistently outperforms a support type-pretrained policy and a randomly-initialized policy, and also seems to outperform Reptile to varying degrees as well. The results also hint at differences in the environments dynamics. For example, entities with evaluation types $1$ or $2$ demonstrate instances where jointly trained and models pretrained on different environments initialize in poor states, suggesting that they require heavier personalization for success. On the other hand, evaluation type $3$ represents a scenario where entity types may be representative of the larger population. While the matched pretrained policy serves as an upper bound in performance for all cases, we observe that the joint trained and unmatched pretrained policy model also seem to effectively transfer to the new environment. 

Lastly we acknowledge the potential challenges of pretraining to personalized environments. Observing the ending coordinates of various policies in Figure 4, we note that in certain situations, such as when the query environment is not representative of a policy's past experiences, methods such as joint training or pretraining with another environment may initiate detrimental behavior such as moving in the wrong direction.

\section{Discussion and Future Work}
We tackle meta-learning for fast adaptation in personalized environments, where environments may share the same reward, but differ dramatically in state transition probabilities. Using ideas from characterizing policy divergence, we introduce
% First introducing a 2D navigation setting for experimental evaluation, we also seek to characterize policy divergence, showing that estimating a state-action probability distribution based on observed trajectories can distinguish trained policies and recover their training environments. Finally, we use these idea to introduce 
competitive improvements to a promising meta-learning algorithm paradigm, better initializing to a set of optimal policy parameters by better organizing past experiences into relevant clusters. While we believe further baselines and experiments are needed, our results also show preliminary support that meta-learning to minimize total distance between all potential optimal parameters may be sub-optimal when trying to adapt to diverse-enough environments. Our method builds on this original motivating curiosity, and suggests one possible alternative to better approach personalized reinforcement learning. 

\begin{figure}[t]
  \centering
  \begin{subfigure}{.33\columnwidth}
    \centering
    \includegraphics[width=\linewidth]{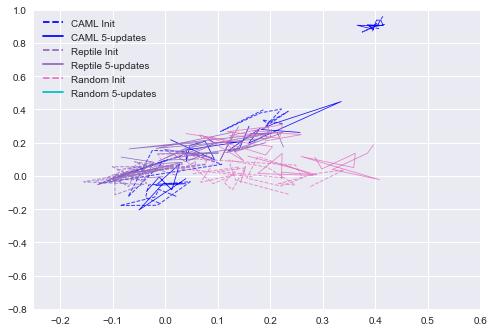}
    % \caption{Agent type 1}
  \end{subfigure}%
  \hfill
  \begin{subfigure}{.33\columnwidth}
    \centering
    \includegraphics[width=\linewidth]{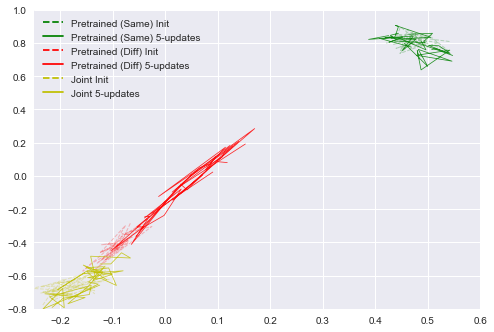}
    % \caption{Agent type 2}
  \end{subfigure}%
  \hfill
  \begin{subfigure}{.33\columnwidth}
    \centering
    \includegraphics[width=\linewidth]{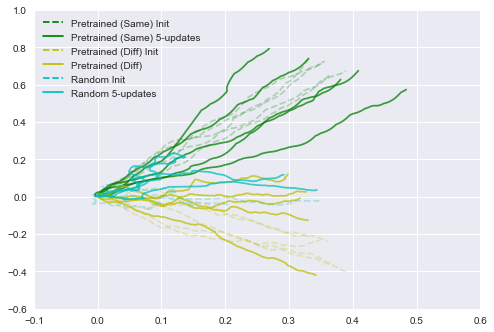}
    % \caption{All agent types}
  \end{subfigure}
  \caption{\textbf{Performance on 2D navigation.} Qualitative comparisons marking the ending location of particles at the end of episodes for the first evaluated type, with attention to performance after initialization and a few updates. Only the matched pretrained model and CAML reach positions noticeably closer to the target coordinate $(1, 1)$ in $5$ updates (left). Visualization of example trajectories taken by pretrained and random policies (right).}
  \vspace{-5pt}
\end{figure}

\newpage 

\bibliographystyle{plainnat}
\bibliography{main}

\end{document}